**SIoU Loss: More Powerful Learning for Bounding Box Regression**
**Zhora Gevorgyan**

**Abstract**

The effectiveness of Object Detection, one of the central problems in computer vision tasks, highly depends on the definition of the loss function — a measure of how accurately your ML model can predict the expected outcome. Conventional object detection loss functions depend on aggregation of metrics of bounding box regression such as the distance, overlap area and aspect ratio of the predicted and ground truth boxes (i.e. GIoU, CIoU, ICIoU etc). However, none of the methods proposed and used to date considers the direction of the mismatch between the desired ground box and the predicted, "experimental" box. This shortage results in slower and less effective convergence as the predicted box can "wander around" during the training process and eventually end up producing a worse model. In this paper a new loss function SIoU was suggested, where penalty metrics were redefined considering the angle of the vector between the desired regression. Applied to conventional Neural Networks and datasets it is shown that SIoU improves both the speed of training and the accuracy of the inference. The effectiveness of the proposed loss function was revealed in a number of simulations and tests. In particular, the application of SIoU to the COCO-train/COCO-val results in improvements of +2.4% (mAP@0.5:0.95) and +3.6%(mAP@0.5) over other Loss Functions.

## Introduction

Object detection is one of the key issues in computer vision tasks and as such it received considerable research attention for decades. It is clear that to solve this problem one needs to define the problem in the concepts acceptable in neural networks methodology. Among these concepts the definition of the so-called loss function (LF) plays a major role. The latter serves as a penalty measure that needs to be minimized during the training and ideally lead to matching of the predicted box that outlines the object to the corresponding ground truth box. There are different approaches in defining LF for object detection problems which take into some sort of a combination of the following "mismatch" metrics of the boxes: the distance between the centers of the boxes, the overlap area, and the aspect ratio. Recently Rezatofighi et. al. claimed that the Generalized IoU (GIoU) LF outperforms state-of-the-art object detection methods with other standard LFs. While these approaches positively affected both training process and the final results we believe there is yet a room for drastic improvements. Thus in parallel with conventional metrics used to calculate the penalty for mismatch of ground truth and model-predicted bounding boxes of objects in the image – namely the distance, the shape and the IoU we propose to take into account also the direction of the mismatch. This addition drastically helps the training process as it results in rather quick drift of the prediction box to the nearest axes and consequent approach needs regression of only one coordinate, X or Y. In short addition of angular penalty cost effectively reduces total number of degrees of freedom.

## Methods

Let's define the metrics that should contribute to SCYLLA-IoU (SIoU) loss function estimation. SIoU loss function consists of 4 cost functions:

- Angle cost
- Distance cost
- Shape cost
- IoU cost

**Angle cost**

The idea behind the addition of this angle-aware LF component is to minimize the number of variables in distance-related "wondering". Basically, the model will try to bring the prediction to X or Y axes first (whichever is closest) and then continue the approach along the relevant axes. To achieve this the process of convergence will first try to minimize $\alpha$ if $\alpha \leq \frac{\pi}{4}$ otherwise minimize $\beta = \frac{\pi}{2} - \alpha$.

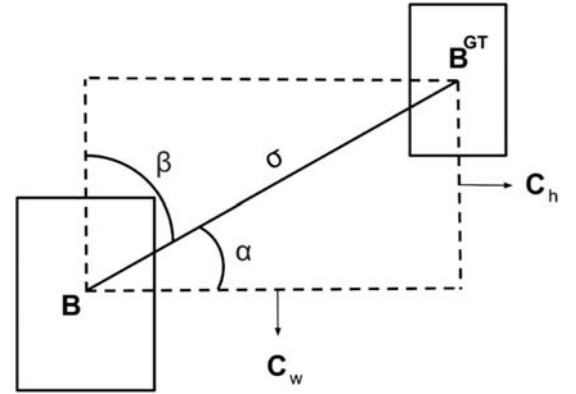

***Figure 1.*** The scheme for calculation of angle cost contribution into the loss function.

To achieve this first, the LF component was introduced and defined in the following way:

$$\Lambda = 1 - 2 * sin^2\left(arcsin\,(x) - \frac{\pi}{4}\right),$$

where

$$x = \frac{c_h}{\sigma} = sin(\alpha)$$

$$\sigma = \sqrt{\left(b_{c_x}^{gt} - b_{c_x}\right)^2 + \left(b_{c_y}^{gt} - b_{c_y}\right)^2}$$

$$c_h = max\left(b_{c_y}^{gt}, b_{c_y}\right) - min\left(b_{c_y}^{gt}, b_{c_y}\right)$$

The curve of the angle cost is displayed on Figure 2.

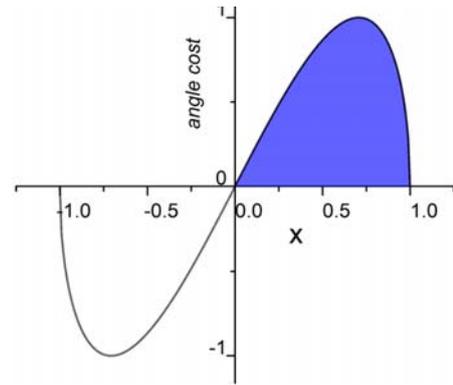

***Figure 2.*** The curve representing the angle cost relative angle ($\frac{\alpha}{2}$)

**Distance cost**

The distance cost is redefined taking into account the angle cost defined above:

$$\Delta = \sum_{t=x,y}(1 - e^{-\gamma \rho_t}),$$

where

$$\rho_x = \left(\frac{b_{c_x}^{gt} - b_{c_x}}{c_w}\right)^2, \rho_y = \left(\frac{b_{c_y}^{gt} - b_{c_y}}{c_h}\right)^2, \gamma = 2 - \Lambda$$

One can see that when $\alpha \to 0$, the contribution of distance cost is drastically reduced. On the contrary – the closer $\alpha$ is to $\frac{\pi}{4}$, the larger is $\Delta$, the contribution. The problem gets harder as the angle increases. So the $\gamma$ was given time priority to the distance value as the angle increases. Note that the cost of distance will become conventional when $\alpha \to 0$.

**Shape cost**

The shape cost is defined as:

$$\Omega = \sum_{t=w,h} (1 - e^{-\omega_t})^{\theta}$$

where

$$\omega_w = \frac{|w - w^{gt}|}{max(w,w^{gt})}, \; \omega_h = \frac{|h - h^{gt}|}{max(h,h^{gt})}$$

and the value of $\theta$ defines how much the shape costs and its value is unique for each dataset. The value of $\theta$ is a very important term in this equation, it controls how much attention should be paid to the cost of the shape. If value of $\theta$ is set to be 1, it will immediately optimize the shape, thus harm free movement of a shape. To calculate value of $\theta$ the genetic algorithm is used for each dataset, experimentally the value of $\theta$ near to 4 and the range that the author defined for this parameter is from 2 to 6.

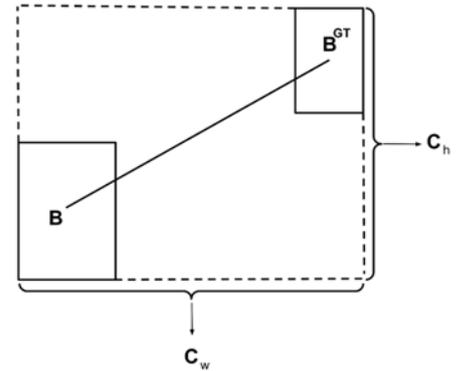

*Figure 3.* Scheme for calculation of the distance between the ground truth bounding box and the prediction of it.

Finally let's define loss function

$$L_{box} = 1 - IoU + \frac{\Delta + \Omega}{2}$$

where

$$IoU = \frac{|B \cap B^{GT}|}{|B \cup B^{GT}|}$$

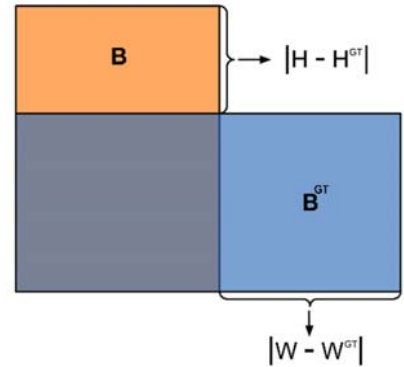

*Figure 4.* Schematic of relation of IoU component contribution.

**Training**

To assess the effectiveness of the proposed loss function the model was trained on COCO dataset - a dataset with 200+K images labeled with 1.5 million object instances. To compare the training effectivity, we performed training of 300 epochs COCO-train using proposed SIoU and the state-of-the art CIoU loss functions and tested on COCO-val set.

**Simulation Experiment**

A simulation experiment was used to further evaluate regression process as proposed by [CIoU paper]. In the simulation experiments, most of the relationships between bounding boxes were covered in terms of distance, scale and aspect ratio. In particular, 7 unit boxes were chosen (i.e., the area of each box is 1) with different aspect ratios (i.e., 1:4, 1:3, 1:2, 1:1, 2:1, 3:1 and 4:1) as target boxes. Without loss of generality, the central points of the 7 target boxes are fixed at (10, 10). The anchor boxes are uniformly scattered at 5000 points (see **Figure 5**). (i) <u>Distance</u>: 5000 points are uniformly chosen within the circular region centered at (10, 10) with radius 3 to place anchor boxes with 7 scales and 7 aspect ratios. In these cases, overlapping and non-overlapping boxes are included. (ii) <u>Scale</u>: For each point, the areas of anchor boxes are set as 0.5, 0.67, 0.75, 1, 1.33, 1.5 and 2. (iii) <u>Aspect ratio</u>: For a given point and scale, 7 aspect ratios are adopted, i.e., following the same setting with target boxes. All the $5000 \times 7 \times 7$ anchor boxes should be fitted to each target box. To sum up, there are in total $1715000 = 7 \times 7 \times 7 \times 5000$ regression cases.

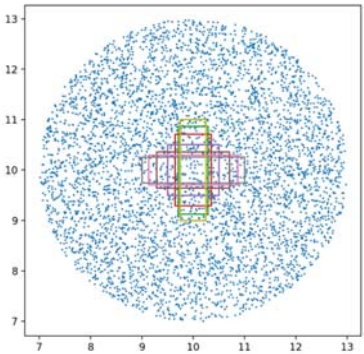

*Figure 5*. Summary image of simulated 5000 anchor boxes centered around (10,10).

Total final error is defined in the following way:

$$E(i) = \sum_{n=0}^{5000} \sum_{t=x,y,w,h} \left| B_t^n - B_t^{GT_n} \right|,$$

where $B^n$ is the current box and $B^{GT_n}$ is the corresponding ground truth box and $E(i)$ is the i-th iteration error.

Adam optimizer with a step learning rate scheduler was used for the training. The initial learning rate and the step size for the step learning were set to 0.1 and to 80, respectively. Training continued for 100 epochs.

**Implementation test**

The final loss function consists of two terms: classification loss and box loss.

$$L = W_{box} \, L_{box} + W_{cls} \, L_{cls}$$

where $L_{cls}$ is focal loss, and $W_{box}, W_{cls}$ are box and classification loss weights respectively. To calculate $W_{box}, W_{cls}, \theta$ a genetic algorithm was used. To train the genetic algorithm a small subset was extracted from the training set and the values were calculated until the fitness value was found less than the threshold or the process exceeded the maximum allowed iterations.

**Results and Discussion**

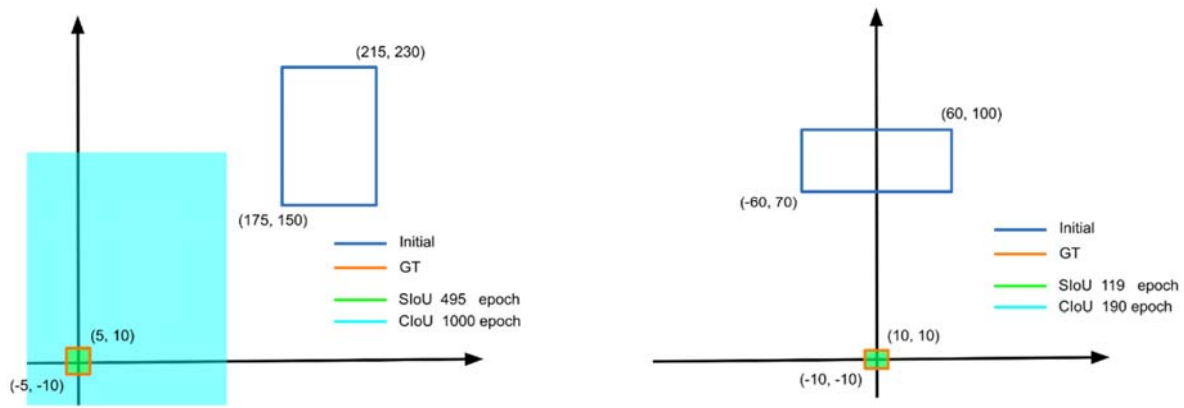

*Figure 6.* Example of simulation showing the convergence of boxes that are placed on axes versus the boxes that are further from axes. Clearly SIoU approach.

One a simple test we compared two cases of convergence – when the initial box is placed on one of the axes (see Fig. 6, right pane) and when the box is away from the axes (see Fig. 6, left pane). Apparently, the advantage of SIoU-controlled training becomes more obvious in cases when the initial prediction box is away from X/Y axes from the ground truth box: SIoU training converges to the ground truth in 495 epochs while the conventional CIoU is not finding it even in 1000 epochs.

Fig. 7 contains the graphs of simulation experiments for CIoU and SIoU. All of the 1715000 regression cases are summed up in 3D plot where X and Y axes are coordinates of the box center point and Z is the error. As you can see, the maximum error with the proposed SIoU loss is almost two orders of magnitude less than with CIoU. Also note that the surface of error is much smoother in case of SIoU which shows that the total error for SIoU is minimal for all simulated cases.

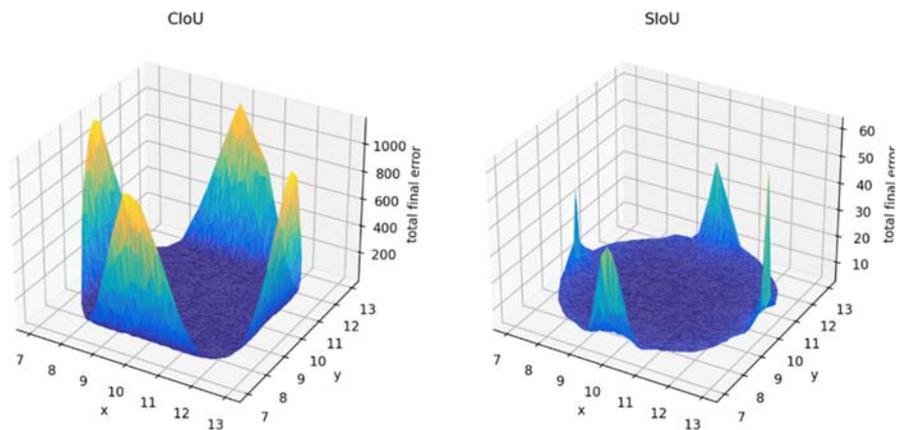

***Figure 7.*** Surface plot of total error for 1715000 simulation cases for CIoU and SIoU.

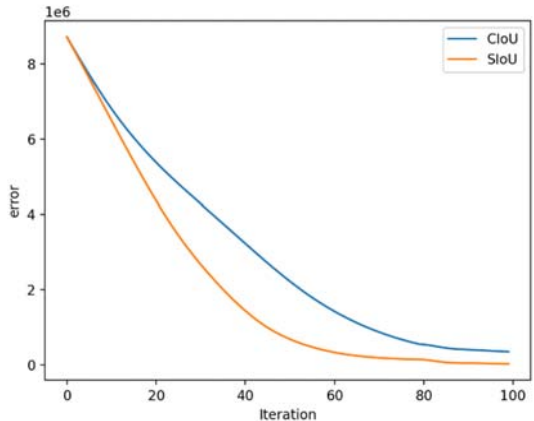

***Figure 8.*** Plot of the errors from CIoU and SIoU losses through the training iterations.

Another result of the comparison of CIoU- and SIoU-powered training is presented on Fig. 8. The dependence of total error on iteration is much steeper for SIoU as well as the final value is lower.

To assess the efficiency of SIoU we also compared its effect on our proprietary Scylla-Net Neural Network. Scylla-Net is a convolution-based neural network that uses genetic algorithm to define its architecture for a specific dataset given predefined layer types. In analogue with different sizes of Darknet models we used two sizes of the model small: Scylla-Net-S and large: Scylla-Net-L.

For a full-feature test we trained the model and monitored all parameters for 300 epochs of training. The corresponding plots are presented in Fig. 9. Clearly all the monitored metrics not only drastically better during the training but also reach better final values.

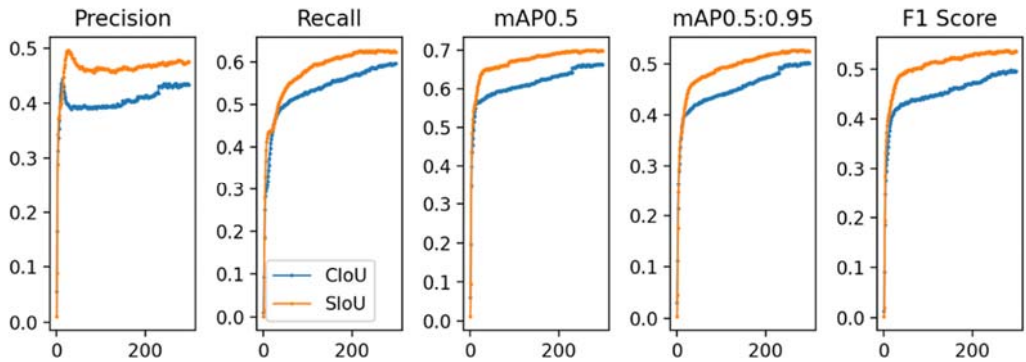

*Figure 9.* Monitored parameters during the training on COCO-train dataset using proposed SIoU and widely used CIoU loss functions.

Namely, mAP with the loss function on COCO-val is 52.7%  mAP@0.5:0.95 (7.6ms including preprocessing, inference and postprocessing) and 70% mAP@0.5 meanwhile with CIoU loss it is only 50.3% and 66.4% respectively. The bigger model can achieve 57.1% mAP@0.5:0.95 (12ms including preprocessing, inference and postprocessing) and 74.3% mAP@0.5 while other architectures like Efficient-Det-d7x, YOLO-V4 and YOLO-V5 can achieve maximum mAP@0.5:0.95 of 54.4% (153ms), 47.1% (26.3ms) and 50.4% (6.1ms tests done with fp16) respectively. Note that YOLO-V5x6-TTA can achieve about 55% on COCO-val but inference time is very slow (~72ms in float precision 16). Fig. 10 summarizes inference times of different models vs. mAP@0.5:0.95. Obviously, Scylla-Net exserts high mAP values while the model inference time is much lower than those of comparisons.

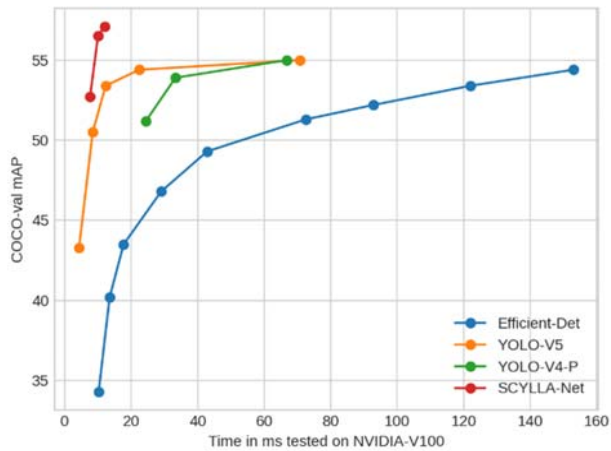

***Figure 10.*** Per-image inference times for different models in relation to mAP@0.5:0.95.

**Table 1.** Comparison of mAP metrics for Scylla-Net trained with CIoU loss, SIoU and SIoU applied to larger Scylla model.

| Network/Loss | mAP@0.5 | mAP@0.5:0.95 |
|---|---|---|
| **Scylla-Net-S/CIoU** | **66.4%** | **50.3%** |
| **Scylla-Net-S/SIoU** | **70.0%** | **52.7%** |
| **Scylla-Net-L/SIoU** | **74.3%** | **57.1%** |

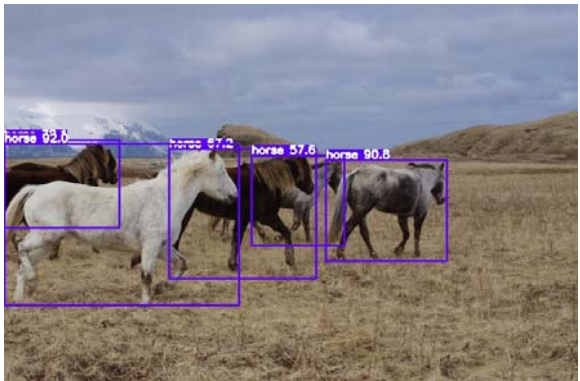 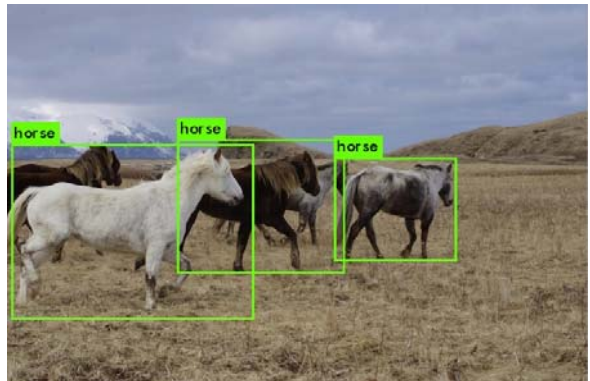

Scylla-Net + SIoU                           YOLO V3

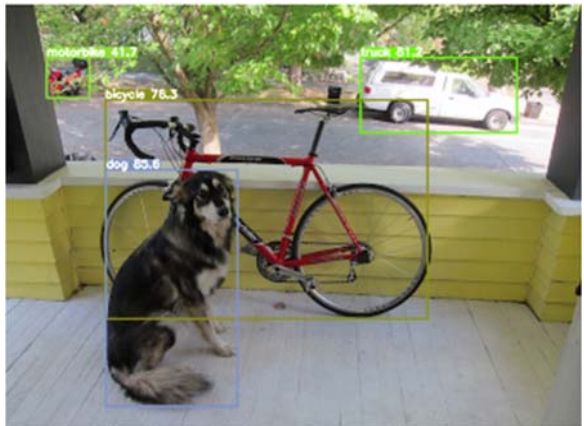 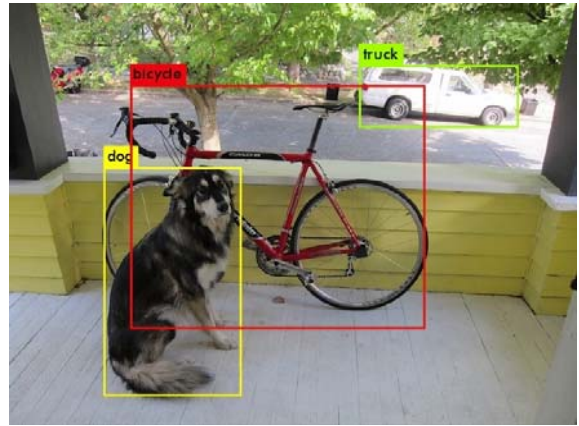

Scylla-Net + SIoU                           YOLO V3

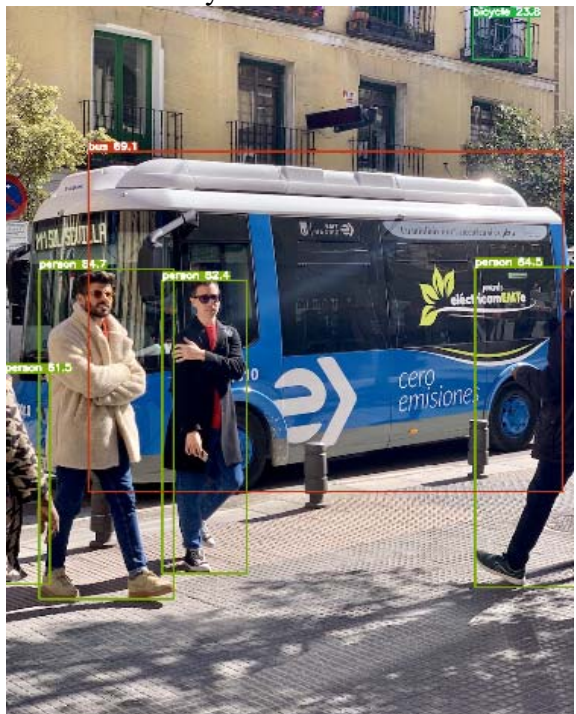 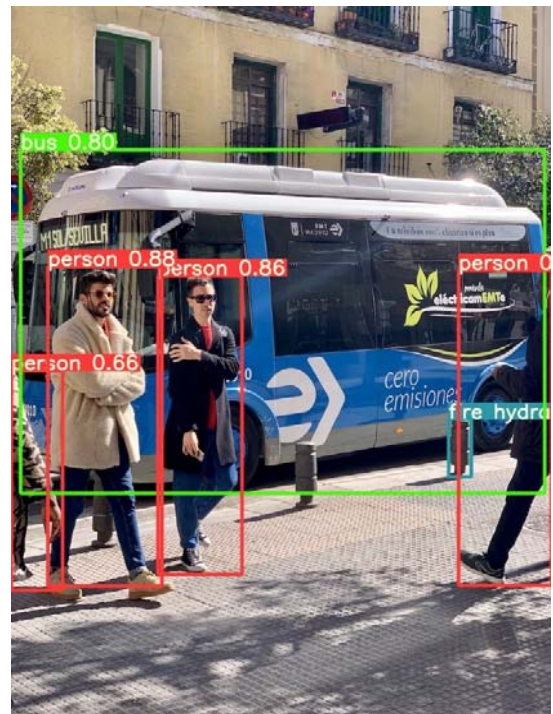

Scylla-Net + SIoU                           YOLO V5

*Figure 11.* Comparison of detections by Scylla-Net + SIoU and other models.

Finally, to evaluate the improvements on model performance we run the sample images presented by different models/approaches through Scylla-Net trained using SIoU. Figure 11 presents some examples. Note the False Negatives from compared models and differences of reported probabilities.

## Conclusion

In this paper, a new loss function for bounding box regression was proposed that can drastically improve both training and inference of Object Detection algorithms. By introducing directionality in the loss function cost, the faster convergence was achieved in the stage of training and better performance in inference compared to the existing methods (such as CIoU loss). Effectively the proposed improvement results in reduction of degrees of freedom (one coordinate vs. two) and the convergence is faster and more accurate.

The claims were verified in comparison with widely used state-of-the art methodologies and measured improvements reported. The proposed loss function can be easily included in any object detection pipeline and will help to achieve superior results.